\documentclass[runningheads]{llncs}
\usepackage{graphicx}
\usepackage{amsmath,amssymb} 
\usepackage{color}
\usepackage[width=122mm,left=12mm,paperwidth=146mm,height=193mm,top=12mm,paperheight=217mm]{geometry}

\usepackage{fancyhdr} 
\usepackage{setspace}

\begin{document}

\title{Generating Shared Latent Variables for Robots to Imitate Human Movements and  Understand their Physical Limitations} 

\titlerunning{Generating Shared Latent Variables for Robots to Imitate Human} 


\authorrunning{M. Devanne and S. M. Nguyen, ECCV Workshops, 2018}

\author{Maxime Devanne and Sao Mai Nguyen \\ \url{nguyensmai@gmail.com} }
\institute{IMT Atlantique, Lab-STICC, UBL}

\maketitle
\thispagestyle{fancy}
\lhead{}
\chead{
\texttt{\begin{spacing}{0.6}
\scriptsize{M. Devanne and S. M. Nguyen. Generating shared latent variables for robots to imitate human movements and understand their physical limitations. In L. Leal-Taixe and S. Roth, editors, Computer Vision  ECCV 2018 Workshops, pages 190 197, Cham, 2019. Springer International Publishing.}
\end{spacing}
 }
\vspace{20pt}}
\rhead{}
\cfoot{}

\begin{abstract}
Assistive robotics and particularly robot coaches may be very helpful for rehabilitation healthcare. In this context, we propose a method based on Gaussian Process Latent Variable Model (GP-LVM) to transfer knowledge between a physiotherapist, a robot coach and a patient. Our model is able to map visual human body features to robot data in order to facilitate the robot learning and imitation. In addition, we propose to extend the model to adapt robots' understanding to patient's physical limitations during the assessment of rehabilitation exercises. Experimental evaluation demonstrates promising results for both robot imitation and model adaptation according to the patients' limitations.

\end{abstract}

\section{Introduction}

Low back pain is a leading cause disabling people particularly affecting the elderly, whose proportion in European societies keeps rising, incurring growing concern about healthcare. 50 to 80\% of the world population suffers at a given moment from back pain which makes it in the lead in terms of health problems occurrence frequency~\cite{who2003burden}. To tackle this chronic low back pain, regular physical rehabilitation exercises is considered most effective~\cite{Kent2012T}.

\begin{figure}[]
\vspace{-0.2cm}
\centering
\includegraphics[width=0.4\columnwidth]{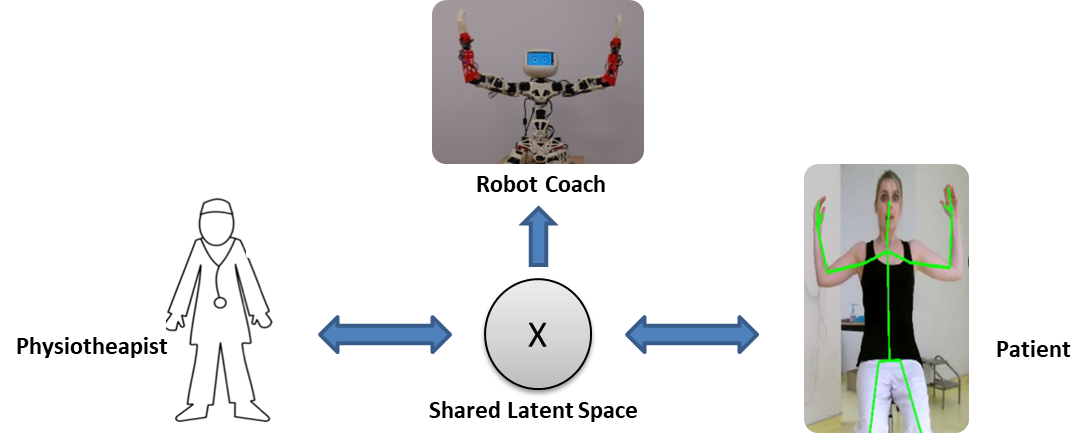}
\includegraphics[width=0.5\columnwidth]{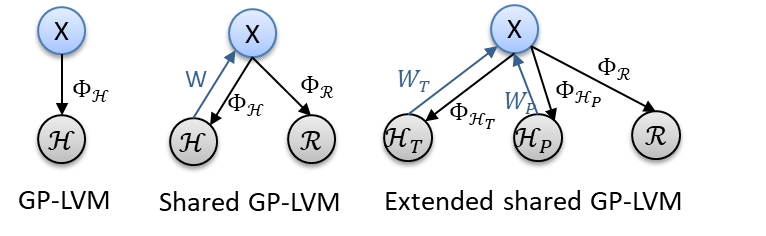}
\caption{\small (Left) Overview of our approach. (Right) Schema of different GP-LVM}\label{overviewSystem}
\vspace{-0.2cm}
\end{figure}

With this perspective, solutions are being developed based on assistive technology and particularly robotics~\cite{Devanne2017,Gorer2017,Devanne2018} where humanoid robots are used for demonstrating rehabilitation exercises to patients. These robots have previously learned these exercises from physiotherapist. However, due to different morphologies between humans and robots, and possible physical limitations of patients, human motion may be difficult to understand by a robot. In this work, we address these issues by training a common low dimensional latent space shared between the therapist, the robot coach and patients, as illustrated in Fig.~\ref{overviewSystem} (left). 
This model allows us to learn an ideal rehabilitation exercise from physiotherapist demonstrations which can be difficult using human data. Moreover, this ideal motion representation is easily interpreted by the robot coach to make it reproduce the correct exercise to the patient. Finally, this model is also employed to adapt the robot's understanding and analysis to the possible physical limitations of patients attending the rehabilitation session.

\section{Related Work}

In the literature, the challenges of robot imitation and motion assessment by robot coaches are usually addressed separately. 

In the context of robot imitation, several vision-based approaches have been proposed. Riley et al.~\cite{Riley2003} proposed  an approach for real-time control of a humanoid by imitation. The imitation is using a stereo vision system to record human trajectories by exploiting color markers on the demonstrators attached to the upper body by inverse kinematics. The authors apply IK to estimate the human’s joint angles and then map it to the robot. Dariush et al.~\cite{Dariush2009} presented an online task space control theoretic retargeting formulation to generate robot joint motions that adhere to the robot’s joint limit constraints, joint velocity constraints and self-collision constraints. The inputs to the proposed method include low dimensional normalized human motion descriptors, detected and tracked using a vision based key-point detection and tracking algorithm. Koenemann et al.~\cite{Koenemann2014} presented a system that enables humanoid robots to imitate complex whole-body motions of humans in real time. The system uses a compact human model and considers the positions of the end effectors as well as the center of mass as the most important aspects to imitate. Stanton et al.~\cite{Stanton2012} used machine learning to train neural networks to map sensor data to joint space. However, these two last approaches employ human motion capture system instead of vision features to capture the human motion. this makes the system not suitable for real-word scenario like physical rehabilitation. 

Only few approaches addressed the challenge of physical rehabilitation through coaching robot systems. While several studies showed the potential of  virtual agents \cite{Waltemate2015PSVRSTV,Anderson2013ACE} and physical robots  \cite{Belpaeme2012JHI}  to enhance engagement and learning in health, physical activity or social contexts, Fasola \textit{et al} \cite{Fasola2013JHI} showed better assessment by the elderly subjects of the physical robot coach compared to virtual systems. Robots  for coaching physical exercises have been recently presented \cite{Goerer2013,schneider2016exercising,Takenori2015}. These approaches employed robots with few degrees of freedom that facilitates the imitation process. However, such robots do not allow realistic movements. 
Moreover, Takenori \textit{et al} \cite{Takenori2015} did not provide any feedback or active guidance to the patient.

In this paper, we employ a humanoid robot with many degrees of freedom called Poppy~\cite{Lapeyre2014} and capture human motion using a kinect sensor with a skeleton tracking algorithm from depth images. We propose a method to simultaneously consider the challenge of robot imitation and human motion assessment in a physical rehabilitation context.

\section{Proposed Approach}
\subsection{Shared Gaussian Process Latent Variable Model}
Our goal is to learn a latent space where we can represent and compare both human and robot poses.
Human upper body poses are characterized by skeletons captured with a kinect sensor providing the 3D position $p_j$ of a set of $J=12$ joints. 
A human pose $y\in \mathcal{H}$ is thus defined as $y = [p_1 \ p_2 \dots p_J]$, where $\mathcal{H}$ denotes the human space. Robot poses are characterized as the motor angles $a_m$ of the Poppy robot including $M=13$ motors.  
Hence, a robot pose $z \in \mathcal{R}$ is defined as $z = [a_1 \ a_2 \dots a_M]$, where $\mathcal{R}$ denotes the robot space. 
To learn such a shared space
, we employ the shared Gaussian Process Latent Variable Model~\cite{Shon2006}

GP-LVM~\cite{Lawrence2004} (See Fig.~\ref{overviewSystem} (right)) is a probabilistic model mapping high dimensional observed data from a low dimensional latent space using a Gaussian process,
 with zero mean and covariance function characterized by a kernel $K$: $f(x) \sim \mathcal{GP}(0,k(x,x'))$. For the kernel $K$, we adopt the popular Radial Basis Function.
The shared GP-LVM is an extension of GP-LVM for multiple data space that shares a common latent space. In our work, we have two observation spaces, the human space $\mathcal{H}$ and the robot space $\mathcal{R}$. 
Given a training set of $N$ human poses $Y = \{y_n\}_{n=1}^N \in \mathcal{H}$ and corresponding robot poses $Z = \{z_n\}_{n=1}^N \in \mathcal{R}$, two mapping functions from the latent space X to observed spaces are defined:
\begin{equation}
f|Y \sim \mathcal{GP}(0,K_Y(X,X'))
\;\;and\;\;
f|Z \sim \mathcal{GP}(0,K_Z(X,X'))
\end{equation}
\noindent where $K_Y$ and $K_Z$ are RBF kernel matrices with hyperparameters $\Phi_Y$ and $\Phi_Z$. In shared GPLVM, optimal latent locations $X^{*}$ are unknown and need to be learned as well as hyperparameters of mappings $\Phi_Y^{*}$ and $\Phi_Z^{*}$. This is done by optimizing the joint marginal likelihood  $p(Y,Z|X,\Phi_Y,\Phi_Z) = p(Y|X,\Phi_Y) \ p(Z|X,\Phi_Z)$. 
We are interesting in mapping data from the human space to robot space through the latent space. Hence, an inverse mapping from the human space to the latent space is required. For that purpose, back constraints are introduced~\cite{Lawrence2006}. 
This feature allows to define latent locations with respect to observed data, $X = h(Y;W)$, where $h$ is an RBF function parameterized by weights $W$. These weights are learned during optimization process instead of latent locations: 
\begin{equation}
\{W^{*}, \Phi_Y^{*}, \Phi_Z^{*}\} = \underset{W,\Phi_Z,\Phi_Z}{\arg\max} \ p(Y,Z|W,\Phi_Y,\Phi_Z)
\end{equation}
As body parts can move concurrently and independently, we consider different shared latent space for each body part separately. Therefore, our approach can also be extended to cases also using lower body parts, by just adding latent spaces for the left and right legs. We use three 2D latent space for the two arms and the spine.
\begin{figure}[]
\vspace{-0.2cm}
\centering
\includegraphics[width=0.25\columnwidth]{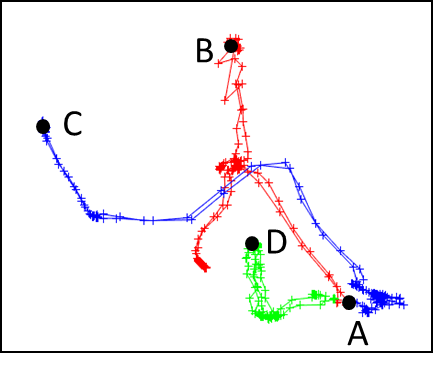}
\;\;\;\;\;\;\;\;\;\;\;\;
\includegraphics[width=0.38\columnwidth]{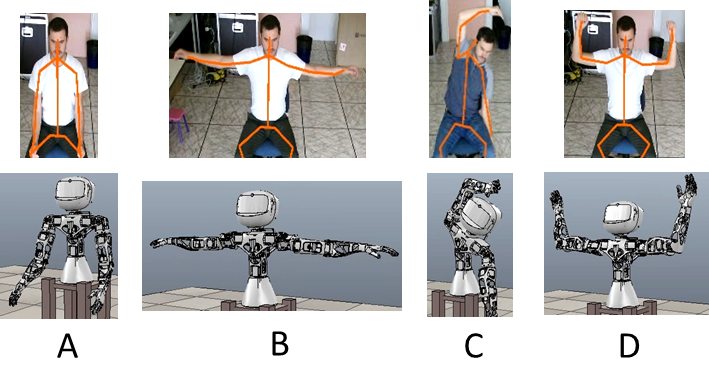}
\caption{\small (left) Three rehabilitation exercises represented in the 2D latent space of the left arm. (right) Corresponding human and robot poses of locations A, B, C and D.}\label{LatenSpace}
\vspace{-0.2cm}
\end{figure}

\subsection{Gaussian Mixture Model on the Latent Space}\label{subsect:GMM}
Once we trained a shared latent space, we can propose to learn a Gaussian Mixture Model on this low dimensional space. This allows to learn an ideal movement from therapist demonstrations projected on the shared space. It can then be employed for robot imitation by projecting back the ideal movement in the robot space.
From $N$ therapist demonstrations $Y^n = [y_1 \ y_2 \dots \ y_T]$, the Gaussian Mixture Model on the latent space is defined as $p(x) = \sum_{k=1}^{K} \phi_k \mathcal{N}(x | \mu_k, \Sigma_k)$, 
where $x$ encodes the human pose $y_t$ projected on the shared latent space.
$K$ is the number of Gaussians, $\phi_k$ is the weight of the $k$-th Gaussian, $\mu_k$ and $\Sigma_k$ are the mean and covariance matrix of the $k$-th Gaussian. The parameters $\phi_k$, $\mu_k$ and $\Sigma_k$ are learned using Expectation-Maximization.
Once a model is learned for each exercise, we generate an optimal sequence using Gaussian Mixture Regression (GMR) which approximates the sequence using a single Gaussian: $p(\hat{x}|t) \approx \mathcal{N}(\hat{\mu}, \hat{\Sigma})$. This optimal sequence is then projected to the robot space to make the robot imitates the expert and demonstrates the exercise to the patient.

\subsection{Transferring Knowledge from Therapist to Patient}\label{Adapation}
In our rehabilitation scenario, the robot coach needs to evaluate the patient's movement
captured using a kinect sensor similarly to therapist's movement.
However, patients needing rehabilitation are often constrained by physical limitations or pain while performing exercises. It may result an incorrect performance even if they did their best to perform the correct exercise. A robust and effective robot coach system must consider such features. We propose to extend the learn shared GP-LVM (see Fig.~\ref{overviewSystem} (right)) by considering two distinct human pose spaces $\mathcal{H}_T$ and $\mathcal{H}_P$ for the therapist and the patient, respectively. $\mathcal{H}_T$ is equivalent to $\mathcal{H}$ described above. $\mathcal{H}_P$ differs from $\mathcal{H}_T$ in the inverse mapping function to the latent space. Specifically, a therapist pose $y_T \in \mathcal{H}_T$ and the corresponding patient pose with physical limitations $y_P \in \mathcal{H}_P$ must be represented by the same point $x$ in the latent space. For that, the weight matrix $W_P$ of the inverse mapping is updated according to the patient. Let $Y_p$ be a patient's performance of an exercise and $X^{*}$ the corresponding ideal demonstration of the same exercise projected on the latent space. The optimization becomes:
\begin{equation}
\{W_P^{*}\} = \underset{W_P}{\arg\max} \ p(Y_p|X^{*},\Phi_Y)
\end{equation}

\begin{figure}[!th]
\vspace{-0.2cm}
\centering
\includegraphics[width=0.25\columnwidth]{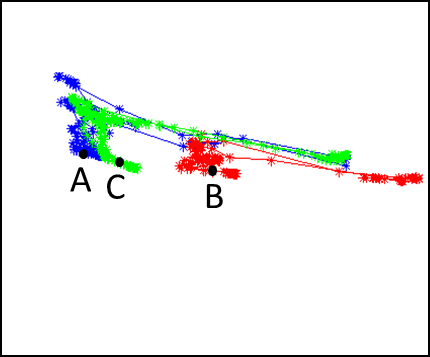}
\;\;\;\;\;\;\;\;\;\;\;\;
\includegraphics[width=0.32\columnwidth]{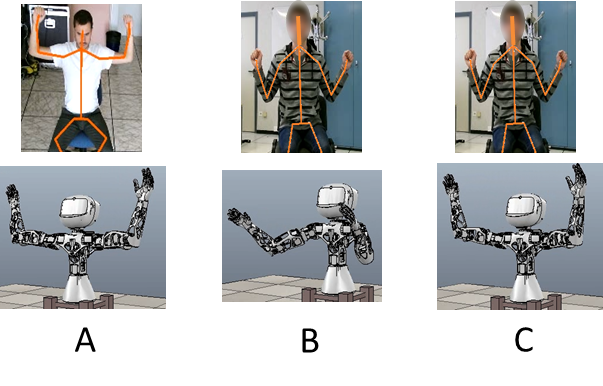}
\caption{\small (left) A wrong exercise in the latent space before (red) and after (green) the model updating. (right) Corresponding human and robot poses of points A, B, and C.
}\label{latentUpdated}
\vspace{-0.2cm}
\end{figure}

The patient specific weight matrix is optimized using gradient descent algorithm. Fig.~\ref{latentUpdated} shows a patients' sequence in the latent space before (red) and after the update (green) in comparison to the ideal therapists' sequence (blue).

\section{Experimental Results}
We evaluate our method on the three rehabilitation exercises selected in cooperation with physiotherapists and performed by two subjects three times~\footnote{Videos are available on www.keraal.enstb.org/exercises.html} playing the role of the physiotherapist and the patient, respectively.
In addition,  subjects performs incorrect exercises by simulating errors~\footnote{Videos are available on www.keraal.enstb.org/incorrectexercises.html}. For the first exercise, the arms are not enough raised. For the second exercise, the subject does not tilt the arm and keep it straight. In the third exercise, the arms are not enough raised. 

For robot movements, we build ideal robot movements with the cooperation of a physiotherapist manipulating the robot in order to perform the desired rehabilitation movement while we record angle positions along the motion. We record one ideal movement per exercise. In addition simulated movements with errors described above are also recorded. These robot movements are used during training of the shared GP-LVM as well as ground truth during evaluation.

\subsection{Imitation Evaluation}
We first evaluate the ability of the approach to perform robot imitation. As described in section~\ref{subsect:GMM}, an ideal motion is generated using GMR on the latent space and the GMM model learned from expert demonstrations. 
This ideal motion is then transferred back to the robot space and compare to the ground truth. We compute the average RMSE error of motor angles between sampled sequence and ground truth. Moreover, we also normalized the RMSE by the standard deviation of motor angles for each exercise to compare the RMSE with the robot's motion.
Results are reported for each exercise in Table~\ref{table:imitation}. 

\begin{table}
\begin{center}
\caption{Robot imitation results.}
\label{table:imitation}
\begin{tabular}{lcccc}
\hline\noalign{\smallskip}
 & Exercise 1 & \hspace{0.2cm}Exercise 2 & \hspace{0.2cm}Exercise 3 & \hspace{0.2cm}Mean\\
\noalign{\smallskip}
\hline
\noalign{\smallskip}
RMSE & 7.1 & 6.9 & 6.1 & \hspace{0.2cm}\textbf{6.7}\\
Normalized RMSE & 0.31 & 0.18 & 0.34 & \hspace{0.2cm}\textbf{0.28}\\
\hline
\end{tabular}
\end{center}
\end{table}
We can see that we obtain a mean RMSE of 6.7 degrees corresponding to $4.1\%$ of the total range of Poppy motor angles. In addition, we obtain a normalized RMSE of 0.28 showing that the RMSE error is much lower than the standard deviation of rehabilitation movements, which represents the noise and the variations in the exercise. This validates the proposed model to imitate therapist demonstration with a high similarity accuracy so as to be clearly understood by the patient. 

\subsection{Therapist-Patient Transfer Evaluation}
We then evaluate the ability of our model to transfer knowledge between a therapist and a patient with physical limitations. 
We first project the error sequence in the shared latent space. Then we project back the sequence to the robot space before and after applying weight updating as described in section~\ref{Adapation}. To show the robustness of the approach, we sample ten random sequences from the latent-robot Gaussian process mapping and compute RMSE error in comparison with ground truth. Average RMSE and standard deviation among the ten sampled sequences are computed. For comparison we also compute such RMSE values for correct sequences of the patient. Results are reported in table~\ref{table:transfer1}.
\begin{table}
\begin{center}
\caption{Therapist-Patient transfer results.}
\label{table:transfer1}
\begin{tabular}{lccc}
\hline\noalign{\smallskip}
Exercise type & \hspace{0.2cm}Exercise 1 & \hspace{0.2cm}Exercise 2 & \hspace{0.2cm}Exercise 3\\
\noalign{\smallskip}
\hline
\noalign{\smallskip}
Correct \hspace{0.5cm} & $8.3\pm0.7$ & $8.0\pm0.9$& $7.4\pm0.8$\\
Incorrect before update & $37.9\pm3.4$ & $17.7\pm1.8$ & $21.9\pm2.4$ \\
Incorrect after update & $14.2\pm1.4$ & $9.1\pm1.0$ & $8.5\pm0.9$\\
\hline
\end{tabular}
\end{center}
\end{table}

We can first observe that, as expected, RMSE errors are much higher for incorrect exercises than for correct exercises. However, if we consider that these errors are due to physical limitations of the patient and apply our updating method, we can see that the RMSE errors becomes close to correct exercises. This means that the robot understands the incorrect exercises similarly to correct exercises. In addition, we propose to deepen the analysis of the third exercise by similarly evaluating a different kind of error (arms are not enough outstretched) with the previously trained model. We obtain RMSE values of $13.4\pm0.89$ and $14.4\pm1.08$ before and after the update, respectively. The similar RMSE values show that by updating the model for one kind of error, it does not affect other type of errors as required in our rehabilitation scenario. 

\section{Conclusions}
We  have proposed a method based on Gaussian Process Latent variable Model for a robot coach system in physical rehabilitation. The method allows to learn a shared space between the therapist and the robot to facilitate robot learning and imitation. The model is then extended to consider variations of patients physical limitations. This allows the robot to understand and assess the patient independently of his physical limitation. Experimental evaluation demonstrates the efficiency of our approach for both robot imitation and model adaptation. 

In the future, we plan to extend our experimental evaluation with more data acquired in real-world environment. Moreover, we would like to investigate the use of key poses instead of full motion sequences during the model training. It would be suitable for a real-world rehabilitation scenario.

\section{Acknowledgement}
The  research  work  presented  in  this  paper  is  partially  supported  by  the  EU
FP7  grant  ECHORD++  KERAAL,  by  the  the  European  Regional  Fund
(FEDER) via the VITAAL Contrat Plan Etat Region and by project AMUSAAL funded by Region Brittany, France.
 
%
%
%
\bibliographystyle{splncs04}
\bibliography{taskcv}

\begin{thebibliography}{10}
\providecommand{\url}[1]{\texttt{#1}}
\providecommand{\urlprefix}{URL }
\providecommand{\doi}[1]{https://doi.org/#1}

\bibitem{who2003burden}
WHO Scientific Group on the Burden of Musculoskeletal Conditions at the Start
  of the New Millennium and others. World Health Organization technical report
  series  \textbf{919}, ~i (2003)

\bibitem{Anderson2013ACE}
Anderson, K., Andr{\'e}, E., Baur, T., Bernardini, S., Chollet, M.,
  Chryssafidou, E., Damian, I., Ennis, C., Egges, A., Gebhard, P., et~al.: The
  tardis framework: intelligent virtual agents for social coaching in job
  interviews. In: Advances in computer entertainment, pp. 476--491. Springer
  (2013)

\bibitem{Belpaeme2012JHI}
Belpaeme, T., Baxter, P.E., Read, R., Wood, R., Cuay{\'a}huitl, H., Kiefer, B.,
  Racioppa, S., Kruijff-Korbayov{\'a}, I., Athanasopoulos, G., Enescu, V.,
  et~al.: Multimodal child-robot interaction: Building social bonds. Journal of
  Human-Robot Interaction  \textbf{1}(2),  33--53 (2012)

\bibitem{Dariush2009}
Dariush, B., Gienger, M., Arumbakkam, A., Zhu, Y., Jian, B., FujiMura, K.,
  Goerick, C.: Online transfer of human motion to humanoids. Int. Journal of
  Humanoid Robotics (IJHR)  \textbf{6}(2) (2009)

\bibitem{Devanne2017}
Devanne, M., Nguyen, S.M.: Multi-level motion analysis for physical exercises
  assessment in kinaesthetic rehabilitation. In: IEEE-RAS 17th International
  Conference on Humanoid Robotics (Humanoids) (November 2017)

\bibitem{Devanne2018}
Devanne, M., Nguyen, S.M., Remy-Neris, O., Le~Gals-Garnett, B., Kermarrec, G.,
  Thepaut, A.: A co-design approach for a rehabilitation robot coach for
  physical rehabilitation based on the error classification of motion errors.
  In: Second IEEE International Conference on Robotic Computing (IRC) (January
  2018)

\bibitem{Fasola2013JHI}
Fasola, J., Mataric, M.: A socially assistive robot exercise coach for the
  elderly. Journal of Human-Robot Interaction  \textbf{2}(2),  3--32 (2013)

\bibitem{Goerer2013}
G{\"o}rer, B., Ali~Salah, A., Akm, H.L.: A robotic fitness coach for the
  elderly. In: 4th International Joint Conference, AmI 2013 (December 2013)

\bibitem{Gorer2017}
Gorer, B., Salah, A.A., Akın, H.L.: An autonomous robotic exercise tutor for
  elderly people. Autonomous Robots  \textbf{41}(3),  657--678 (7 2017)

\bibitem{Kent2012T}
Kent, P., Kjaer, P.: The efficacy of targeted interventions for modifiable
  psychosocial risk factors of persistent nonspecific low back pain--a
  systematic review. Manual therapy  \textbf{17}(5),  385--401 (2012)

\bibitem{Koenemann2014}
Koenemann, J., Burget, F., Bennewitz, M.: Real-time imitation of human
  whole-body motions by humanoids. In: IEEE International Conference on
  Robotics and Automation (ICRA) (June 2014)

\bibitem{Lapeyre2014}
Lapeyre, M.: Poppy: open-source, 3D printed and fully-modular robotic platform
  for science, art and education. Ph.D. thesis, Universit{\'e} de Bordeaux
  (2014)

\bibitem{Lawrence2004}
Lawrence, N.D.: Gaussian process latent variable models for visualisation of
  high dimensional data. In: Advances in neural information processing systems
  (December 2006)

\bibitem{Lawrence2006}
Lawrence, N.D., Candela, J.Q.: Local distance preservation in the gp-lvm
  through back constraints. In: International Conference on Machine Leraning
  (ICML) (December 2006)

\bibitem{Takenori2015}
Obo, T., Loo, C.K., Kubota, N.: Imitation learning for daily exercise support
  with robot partner. In: Robot and Human Interactive Communication (RO-MAN),
  2015 24th IEEE International Symposium on. pp. 752--757. IEEE (2015)

\bibitem{Riley2003}
Riley, C., Ude, A., Wade, K., Atkeson, C.: Enabling real-time full-body
  imitation: a natural way of transferring human movement to humanoids. In:
  IEEE International Conference on Robotics and Automation (ICRA) (September
  2003)

\bibitem{schneider2016exercising}
Schneider, S., K{\"u}mmert, F.: Exercising with a humanoid companion is more
  effective than exercising alone. In: Humanoid Robots (Humanoids), 2016
  IEEE-RAS 16th International Conference on. pp. 495--501. IEEE (2016)

\bibitem{Shon2006}
Shon, A., Grochow, K., Hertzmann, A., Rao, R.P.: Learning shared latent
  structure for image synthesis and robotic imitation. In: 18th International
  Conference on Neural Information Processing Systems (December 2006)

\bibitem{Stanton2012}
Stanton, C., Bogdanovych, A., Ratanasena, E.: Teleoperation of a humanoid robot
  using full-body motion capture, example movements, and machine learning. In:
  Australasian Conference on Robotics and Automation (ACRA) (December 2012)

\bibitem{Waltemate2015PSVRSTV}
Waltemate, T., H{\"u}lsmann, F., Pfeiffer, T., Kopp, S., Botsch, M.: Realizing
  a low-latency virtual reality environment for motor learning. In: Proceedings
  of ACM Symposium on Virtual Reality Software and Technology (VRST) (2015)

\end{thebibliography}

\end{document}